\documentclass[journal]{IEEEtran}
\usepackage[utf8]{inputenc}

\usepackage[pdftex]{graphicx}
\usepackage[cmex10]{amsmath}
\usepackage{array}
\usepackage{url}
\usepackage{xcolor}
\usepackage{amsfonts}
\usepackage{multirow}
\usepackage{pdflscape}
\usepackage{afterpage}
\usepackage{capt-of}
\usepackage{textcomp} 

\begin{document}

\title{Interactive segmentation of medical images through fully convolutional neural networks}
\author{ Tomas Sakinis,
         Fausto Milletari,
         Holger Roth,
         Panagiotis Korfiatis,
         Petro Kostandy,
         Kenneth Philbrick,
         Zeynettin Akkus,
         Ziyue Xu,
         Daguang Xu,
         Bradley J. Erickson
\IEEEcompsocitemizethanks{ 
\IEEEcompsocthanksitem Tomas Sakinis is with Oslo University Hospital, Oslo, Norway and Mayo Clinic, Rochester, MN, USA.
\IEEEcompsocthanksitem Fausto Milletari, Holger Roth, Daguang Xu and Ziyue Xu are with Nvidia Corporation, Santa Clara, CA, USA.
\IEEEcompsocthanksitem Panagiotis Korfiatis, Petro Kostandy, Kenneth Philbrick, Zeynettin Akkus and Bradley J. Erickson are with Mayo Clinic, Rochester, MN, USA.
}%
}

\maketitle

\begin{abstract}
Image segmentation plays an essential role in medicine for both diagnostic and interventional tasks. Segmentation approaches are either manual, semi-automated or fully-automated. Manual segmentation offers full control over the quality of the results, but is tedious, time consuming and prone to operator bias. Fully automated methods require no human effort, but often deliver sub-optimal results without providing users with the means to make corrections. Semi-automated approaches keep users in control of the results by providing means for interaction, but the main challenge is to offer a good trade-off between precision and required interaction. 
In this paper we present a deep learning (DL) based semi-automated segmentation approach that aims to be a "smart" interactive tool for region of interest delineation in medical images. We demonstrate its use for segmenting multiple organs on computed tomography (CT) of the abdomen.
Our approach solves some of the most pressing clinical challenges: (i) it requires only one to a few user clicks to deliver excellent 2D segmentations in a fast and reliable fashion; (ii) it can generalize to previously unseen structures and "corner cases"; (iii) it delivers results that can be corrected quickly in a smart and intuitive way up to an arbitrary degree of precision chosen by the user and (iv) ensures high accuracy. We present our approach and compare it to other techniques and previous work to show the advantages brought by our method. 

\end{abstract}


\IEEEpeerreviewmaketitle

\section{Introduction}

Image segmentation has become an increasingly important task in radiology research and clinical practice. Quantitative analysis of medical data often requires precise delineation of organs, abnormalities, and structures of interest which can be delivered by computer assisted approaches as demonstrated by recent literature \cite{giger2008anniversary,iglesias2015atlas,peng2013graph,olabarriaga2001interactive,petitjean2011cardiac,nobel2006ultrasound,foster2014pet}.

Current segmentation approaches can be grouped in three classes: Manual, semi-automatic and fully-automatic. 
\begin{figure} 	
\centering 	
\includegraphics[scale=0.24]{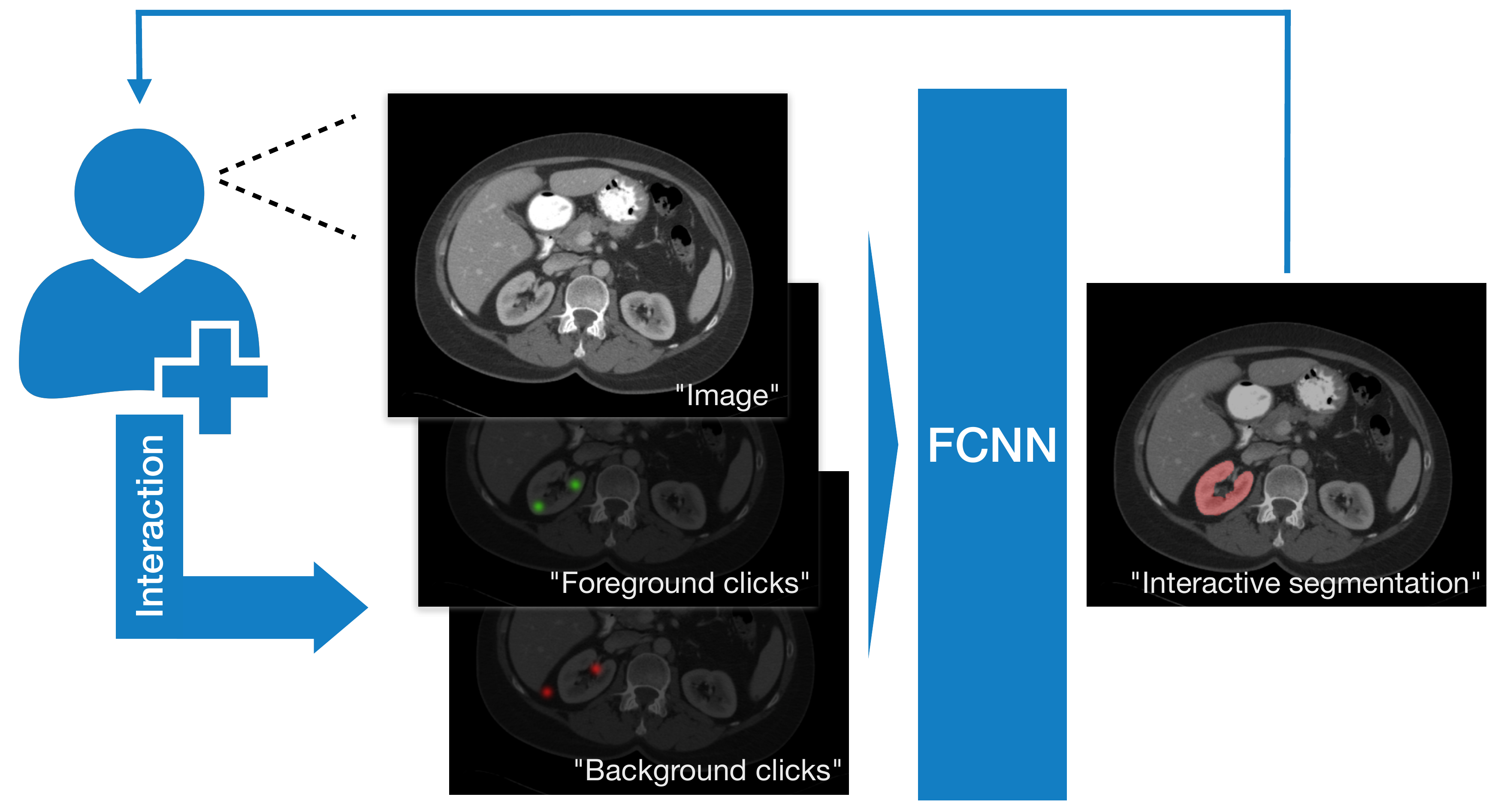} 	
\caption{Schematic representation of our approach. The user, who possesses domain knowledge, observes the image and clicks on the region of interest (green Gaussians). Almost instantly the 2D segmentation is generated. If the result needs correction, the user can drag the green Gaussian around to see live as the segmentation results change. If that is not adequate, the user can place any number of additional clicks inside (green) or outside (red) of the region of interest to steer the DL-based segmentation according to his or her requirement and preferences. The user keeps interacting with the segmentation results until segmentation can be considered correct (usually 1 to 3 clicks).} \label{fig:summary} 
\end{figure}

Manual segmentation approaches are tedious, time consuming and can be affected by inter- and intra- observer variability \cite{leong2019segmentation}. Each pixel of the image needs to be manually assigned to its class and, although very accurate results can be obtained with this technique, the time required challenges the translation of such algorithms into clinical practice. For some tasks manual segmentation can require hours for a single case \cite{kline2017performance}. 

Fully-automatic methods do not require user interaction. Earlier approaches leverage a wide range of techniques. Simpler methods, such as thresholding, have a very short run-time but rarely deliver satisfactory results, while other more complex algorithms, such as atlas based techniques, can only deliver results after a few hours of computation. More recent approaches use DL \cite{ronneberger2015u,milletari2016v,cciccek20163d} to produce fast segmentations without any user interaction. Although these approaches have demonstrated state-of-the-art performances on a wide range of tasks and on public datasets, they still find limited applicability in practical scenarios. There are multiple possible reasons for this: (i) automatic methods are sensible to small domain shifts between original data used for training and supplied new data, often resulting in failure; (ii) it is not easy to recover from these failures because there is no mechanism of user interaction for smart correction of possible mistakes; (iii) there is typically no way to adapt a fully-automatic approach to a specific patient or population by interacting with the segmentation results; (iv) it is typically not possible to generalize the approach to structures that were not observed during training; and (v) the lack of an interaction mechanism makes it hard to understand the failure modes of the algorithm leading to a drop in user confidence. 

Semi-automatic approaches \cite{an2017accuracy} are already widespread and used in medicine. They are integrated in popular publicly available software packages such as ITK-Snap \cite{itksnap}, 3D Slicer \cite{3dslicer} and other similar tools \cite{kline2016semiautomated}. These tools allow users to obtain segmentations usually through a fairly intensive annotation and correction process. Recently, DL-based approaches have been employed for semi-automatic annotation in the field of computer vision. The results shown in \cite{wang2018interactive}, \cite{xu2016deep} demonstrate that the users of these algorithms can obtain good segmentation results through simpler forms of interaction, such as scribbles and clicks.

In this paper we present a semi-automatic segmentation method that leverages convolutional neural networks (CNNs). Trained on fairly limited training sets, it is capable of producing fast and accurate segmentation outputs on 2D CT images. The algorithm (i) responds to a simple form of interaction by the user, more specifically mouse clicks; (ii) is able to deliver accurate results with the smallest amount of interaction (1 click), but allows for refinement of the results up to arbitrary precision by moving the click position or adding additional interactions. In other words, as the number of interactions grows, performance grows accordingly; (iii) generalizes and performs well on previously unseen structures.

We foresee application of our approach for both purely clinical workflows that require segmentation, and to data annotation tasks that aim to produce large amounts of annotated images for (re-)training/testing of computer assisted segmentation methods.
We support our claims by extensive experimental evaluation as discussed in Section \ref{section:experiments}.

\section{Previous work}

Segmentation of medical images has a wide range of applications. One of the primary uses of segmentation in radiology is for making quantitative measurements, more specifically, measuring the area and volume of segmented structures. Volumetric analysis has been shown to be more robust than caliper based distance measurements which are most commonly used due to simplicity and a historical lack of well established segmentation methods \cite{devaraj2017nodulevolume}. Some other uses of segmentation include visualization, 3D printing, radiation treatment planning and are often an integral step in other applications such as registration, radiomics and image guided surgery. Some of these use cases are fairly new and used increasingly. As the segmentation tools keep improving, we believe that it is safe to assume that the use of medical image segmentation will keep increasing in both research and clinical practice.


Semi-automatic segmentation approaches such as \cite{petitjean2011review} use shallow image features and user interaction in the form of bounding boxes or scribbles to perform segmentation. An early work \cite{udupa1996fuzzy} employed region similarity to establish "objectness" and "connectedness" during segmentation. Graph cut \cite{boykov2006graph}, normalized cuts \cite{shi2000normalized}, geodesics \cite{akkus2015semi} and other graph based methods such as random walks \cite{grady2005random} have been applied to 2D and, in some cases, 3D segmentation. The main finding is that semi-automatic methods can deliver good segmentation when the regions to be segmented are fairly uniform and when their characteristics can be accurately captured by low level features such as edges, intensity differences between adjacent pixels and other local information. Unfortunately, in more complex settings, due to the well-known limitations of shallow features, these approach end up requiring extensive user interaction in order to deliver segmentations of sufficient quality.


More recently, state of the art results have been obtained by DL methods for fully-automatic segmentation. Segmentation in 2D \cite{ronneberger2015u} and 3D \cite{milletari2016v,cciccek20163d} has been obtained using fully convolutional neural networks (FCNNs) with an encoding-decoding architecture. Unfortunately, although these methods and their successors \cite{menze2015multimodal,roth2018multi,larsson2016deepseg} have been demonstrated to deliver state of the art performances on publicly available and very challenging datasets, they offer no means for user interaction. This aspect limits applicability in clinical settings, where precision tends to be prioritized over the speed that these automatic methods usually grant. 


The computer vision community has recently proposed various novel approaches to tackle semi-automatic segmentation through DL methods \cite{xu2016deep,maninis2017deep,agustsson2018interactive}. 
Semi-automatic approaches have cut the time requirements (measured in hours) for annotation by a factor of 10 \cite{kline2017performance}. Moreover it has been found that DL based semi-automatic methods can deliver accurate results while providing a natural way for the users to interact and possibly to "understand" the predictions of the network.
Instead of relying on shallow features, as earlier approaches would do, deep convolutional neural networks are able to capture complex patterns and learn the notion of "objectness" \cite{xu2016deep} resulting in superior results. 

User interaction can be supplied in different ways. In \cite{castrejon2017annotating} users needed to supply bounding boxes around the objects of interest to guide segmentation; other works have made use of scribbles \cite{rother2004grabcut,grady2005random,boykov2006graph}, "extreme points" \cite{maninis2017deep,agustsson2018interactive} and clicks \cite{xu2016deep}. 


Interactive segmentation methods in medical imaging are based on extensions of the above mentioned methods. Recently, Wang et al. \cite{wang2016slic} used a minimally interactive framework to segment the human placenta in fetal MRI, combining online random forests, conditional random fields, and 4D graph cuts. DeepIGeoS \cite{wang2018deepigeos} uses DL and feeds geodesic distance transforms of user scribbles as additional CNN channels to allow interactive segmentation. However, segmentation performance might deteriorate for unseen object classes. As an alternative, image-specific fine-tuning has been proposed to be incorporated into bounding box and scribble-based segmentation \cite{wang2018interactive}.
Another recent DL-based method relies on Random Walks to provide an initial scribble-based segmentation for supervision of fully convolutional neural networks (FCNNs) has been been applied to the semi-automatic segmentation of cardiac CT images \cite{can2018learning}.


Motivated by the findings of recent research, to minimize the required interaction, we have decided to employ click-based user interaction for our approach. Differently than \cite{xu2016deep}, which relies on distance maps obtained from foreground and background clicks, we follow a strategy similar to \cite{maninis2017deep} where distance maps have been replaced with Gaussians for superior performance.

Similarly to \cite{xu2016deep} we use model predictions for creation of training data. Our approach differs in that it allows an arbitrary amount of predictions mimicking human behavior for each training data case, not limited to one. Also, in contrast to \cite{xu2016deep}, we use a model based on an encoder-decoder architecture.

\begin{figure*} 	
\centering 	
\includegraphics[scale=0.35]{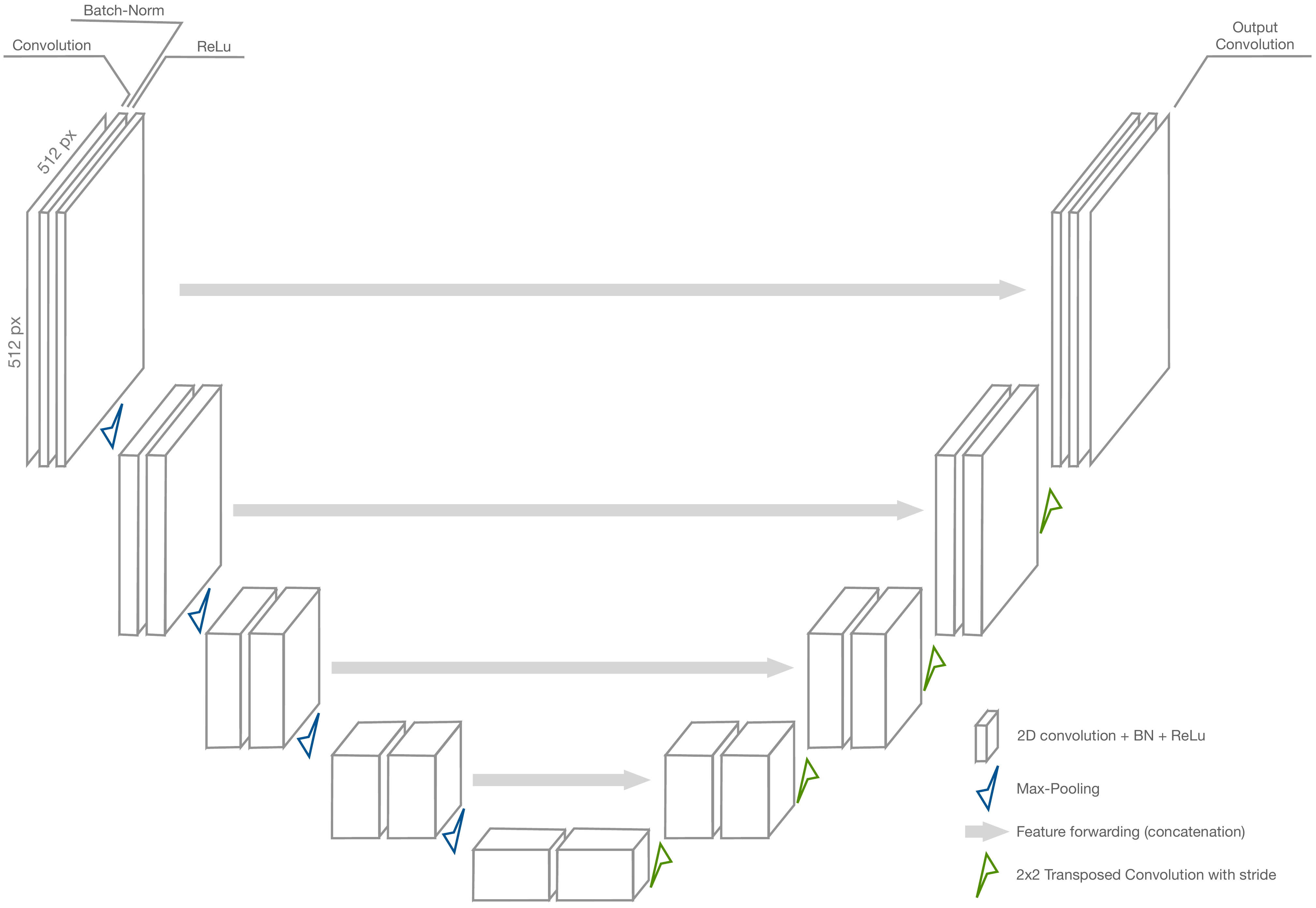} 	
\caption{Schematic representation of our FCNN architecture. The structure of this network was inspired by \cite{ronneberger2015u} and augmented with batch normalization (BN) and convolutions with padding. Each block performs two convolutions followed by batch normalization and ReLu activations.} \label{fig:architecture} 
\end{figure*}

\section{Methods}
Our approach uses a (FCNN) architecture to perform interactive 2D segmentation of medical images. The network is trained to segment only one region of interest at a time, taking into account user input in the form of one or more mouse clicks. Different organs or other discrete structures can thus be delineated based on where in the image these clicks are placed. 

The network is trained using raw 2D images and "guidance signal" as input and a binary mask of one specific segmentation object as output. That is, a 2D image with multi-label ground truth consisting of $K$ different segments results in a set of $K$ images with binary labels and associated guidance signals. 

\subsection{Fully convolutional architecture}

Our network architecture is inspired by \cite{ronneberger2015u}. We have modified the structure to include a down-sampling path with four blocks,
 a bottleneck block and an up-sampling path with four blocks. 
Each block comprises two convolutions, two batch normalization, and two rectified linear units (ReLu) which are interleaved (Figure \ref{fig:architecture}). The down-sampling path makes use of $2 \times 2$ max-pooling operations applied with a stride of $2$ to reduce the resolution and increase the spatial support of the features. The up-sampling path makes use of $2 \times 2$ transposed convolutions applied with a stride of $2$ to increase the size of the feature maps and allow prediction of a full resolution segmentation output.

This choice of network architecture is motivated by recent literature, but in our experience, any architecture comprised of an autoencoder with fully convolutional networks that has been demonstrated being adequate to perform medical image segmentation can be used instead with comparable performance.

\subsection{User interaction and guidance signals}
\label{section:interaction}
Our method relies on user guidance and interactions to deliver accurate segmentation. Users are able to interact with the algorithm via mouse clicks through a user interface (UI) which accepts two types of inputs:

\begin{itemize}
    \item "foreground clicks" - which should be placed within the structure of interest that the user wants to segment; they "guide" the network towards predicting foreground. If the result has false negative areas, additional "foreground clicks" can be placed within false-negative areas to reduce these errors.
    \item "background clicks" - which should be placed wherever the user feels that the current structure of interest is over-segmented in order to correct for those false positive errors; they "guide" the network towards reducing foreground prediction.
\end{itemize}

Interactions are converted to two guidance signals and supplied to the the network. These signals respectively capture "foreground clicks" and "background clicks". They take the form of images with the same spatial size as the input data, and having zero value except for the pixels corresponding to clicks. Those pixels are set to unitary value. The generated images subsequently are smoothed with a Gaussian filter and normalized to $[0,1]$ range. Finally, input data and guidance signals for foreground and background interactions are concatenated and presented to the network as three-channel 2D inputs.

The minimum amount of interaction for a single image is 1-click, which should be placed within the object of interest. In this way, the network receives the necessary guidance to perform an initial segmentation which, although usually accurate (Section \ref{section:experiments}), can be refined by supplying further guidance. Additional foreground and background clicks can be provided until the desired segmentation quality is obtained. 

\subsubsection{Simulated interaction}
During both training and large-scale testing, true human interaction is unfeasible. We therefore simulate user interactions which aim to mimic the way a human expert would use this interactive segmentation method. 

The first interaction, aiming to steer the CNN prediction towards a segmentation of the current structure of interest, is usually placed in the innermost portion of the region of interest. Subsequent interactions, which aim to correct and improve the obtained segmentation, usually correspond to regions of larger and more noticeable errors \cite{agustsson2018interactive}.

In our approach each pixel of the image to be segmented is assigned a probability of being used for interaction which is derived from distance fields obtained from a disparity map
\[
D=Gt - P=
\begin{cases}
+1 $ if $ Gt = 1 $ and $ P = 0,\\
-1 $ if $ Gt = 0 $ and $ P = 1,\\
0 $ otherwise $
\end{cases}
\]
where $Gt$ is the ground truth binary signal and $P$ represents the prediction. Two binary images, $D^+$ and $D^-$, are then obtained by considering only false-negative and false-positive regions of $D$ and Chamfer distance fields between unmarked and marked pixels are obtained. The probability of an image location $x,y$ to be used for guidance towards foreground or background segmentation depends on these distances. In particular, when indicating the distance fields relative to $D^+$ and $D^-$ with $DD^+$ and $DD^-$, we obtain the probabilities (up to a multiplicative constant) of a location $(x,y)$ to be used for a foreground and background interaction as $P^+(x,y) \propto exp(DD^+(x,y)) - 1$ and $P^-(x,y) \propto exp(DD^-(x,y)) - 1$ respectively. In other words, interactions are more likely to occur in correspondence of the innermost portion of larger false-positive/negative regions rather than elsewhere.

The first interaction is produced assuming an initial prediction $P$ that is zero everywhere. This results in a $D^+ \equiv Gt$ and $D^- = 0$. Every subsequent interaction is obtained by supplying the binary predicted contour $P$ as obtained from the forward pass of the CNN, computing the disparity maps and relative distance fields and placing a simulated "click" aiming to correct false-positives or false-negatives. If the area of false positives is higher than the false negatives, a "background click" is simulated, otherwise the algorithm supplies a "foreground click". 

\begin{figure*} 	
\centering 	
\includegraphics[scale=0.305]{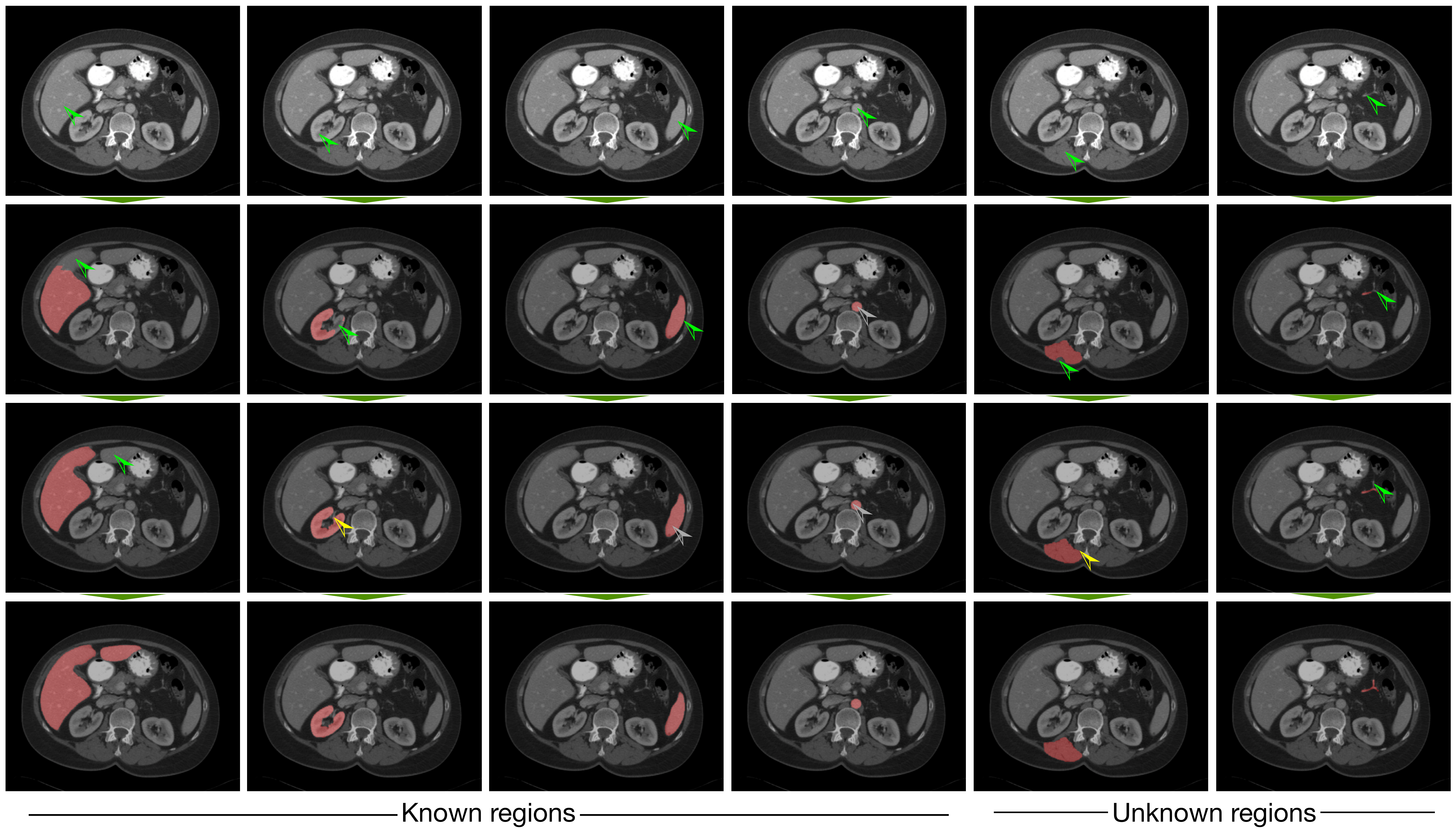} 	
\caption{Example of interactive segmentation. The first four columns show result on learned structures whose class has been observed during training while the last two columns show results on "unknown" structures. Depending on user clicks, represented here with arrows, the segmentation algorithm produces different predictions in real time. Green arrows indicate "foreground clicks", yellow arrows "background clicks", grey arrows indicate interactions which did not have any visually noticeable influence on the results.} \label{fig:interactive_seg} 
\end{figure*}

\subsection{Training strategy}

Our method performs binary segmentation of arbitrary organs and structures by taking into account user interactions. The user interaction steers the network towards delineating a specific foreground region. In our experience any region that shows a distinctive pattern or some form of spatial coherence can be segmented in this way. Some previously unseen regions might require a higher number of interactions before good results can be achieved, but our method has demonstrated the capability of segmenting previously unseen regions during routine usage (Section \ref{section:experiments:unkregions}).

We chose to demonstrate our approach using a multi-region (multi-label) dataset. Each image is associated with multi-label ground truth depicting $S$ segments. We convert this multi-region dataset to a binary dataset. Each image is associated to $S$ different ground truth labels, each referring to a specific organ. 

We supply the network with tuples $\{I,Gt,U_0\}$ consisting respectively of an image $I$, ground truth label $Gt$ and initial guidance signal $U_0$. The guidance signal needs to be consistent with the current foreground label. It initially contains only one "foreground click" placed in correspondence of one of the pixels of the foreground annotation $Gt$.

During training, we iterate across the whole training set batch-by-batch.
Initially, the network is used in "prediction" mode. Each batch is processed $K$ times, with $K$ the maximum number of allowed interactions. User-interaction, obtained as described in Section \ref{section:interaction}, is supplied iteratively according to the observed segmentation errors. That is, for each iteration $k$ beyond the first, new user guidance $U_k$ is obtained by taking into account the quality of the results $P_{k-1}$. The user interaction signal $U_k$ supplied to the network during the $k$-th iteration depends on the previous interaction $\{I,Gt,U_{k-1}\}$ and in particular on the current guidance signal $U_{k-1}$. The network does not learn from these interactions and neither losses or parameters updates are computed in this stage.

In order to force the network to learn to segment accurately with the minimum number of "clicks", the probability of an interaction to take place decreases linearly as $k$ increases. 
Once we have obtained up to $K$ interactions for each image in the batch, we supply the final training tuple $\{I,Gt,U_{t}\}$ to the network. The number of actual interactions, $t$, takes a value from $0$ to $K$ for each of the $N$ images in the batch.
We optimize the parameters of our FCNN using the average Dice loss $L = \sum_{j}^{N}\frac{1 - Dice_j}{N}$ over the $N$ training images in each batch at each iteration. 
The Dice score \cite{milletari2016v} formulation used in our work is  
\[
Dice=\frac{2\sum_{i}^{N}Gt_{i}\cdot
P_{i}}{\sum_{i}^{N}Gt_{i}^{2}+\sum_{i}^{N}P_{i}^{2}}.
\]

\section{Experiments}
\label{section:experiments}

Our method has been trained and validated on one public dataset and tested on two. These datasets contain CT images of the abdomen. They were respectively released during the "multi-atlas labeling: beyond the cranial vault ("BCV" dataset)\footnote{\url{https://www.synapse.org/#!Synapse:syn3193805/wiki/217752}}" MICCAI challenge that was held in 2015 and the "medical segmentation decathlon challenge ("MSD" dataset)\footnote{\url{http://medicaldecathlon.com}}" held at MICCAI 2018 \cite{simpson2019large}. There is no overlap between the datasets. The first dataset includes 30 training CT volumes with corresponding ground truth segmentations and 20 test cases with secret, withheld ground truth segmentations. The second dataset includes both MRI and CT volumes relevant to 10 different segmentation tasks. Out of these tasks we select 2 which use CT data of the abdomen. We perform spleen and colon cancer delineation on this data. By testing on a completely different dataset than the one used for training we show the applicability of our approach to clinical use-cases. 

\begin{figure*} [hbt!]
\centering 	
\includegraphics[scale=0.33]{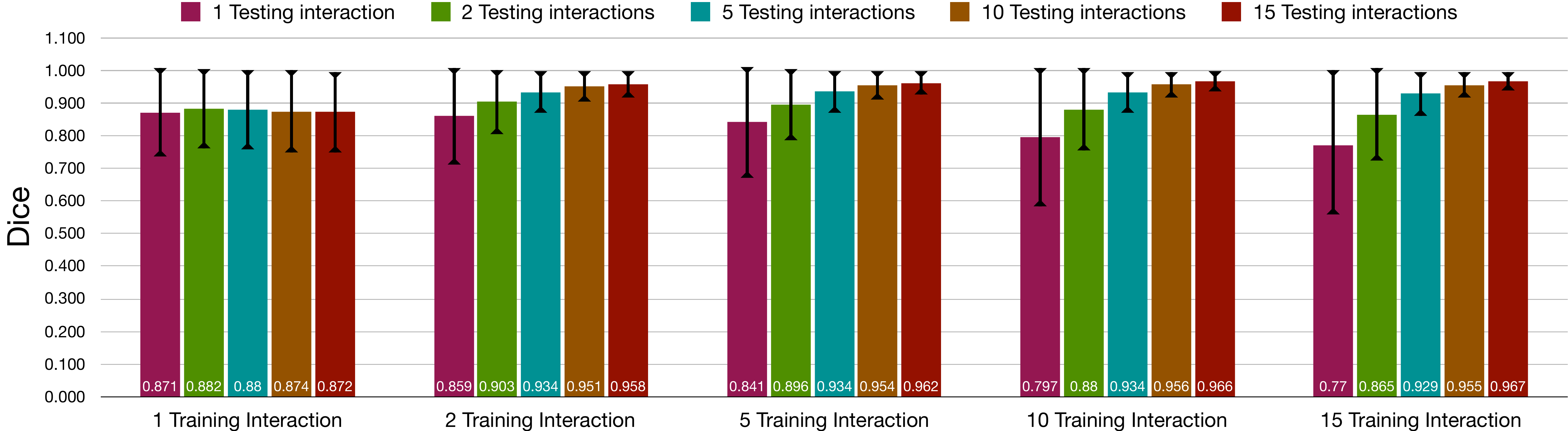} 	
\caption{Average Dice coefficients across cross-validation folds (higher is better). Different amounts of training-time and test-time interaction have been supplied to the algorithm. Groups represent the amount of interactions per slice and organ during training. Colors represent the amount of interactions during testing.} \label{fig:crossVDice} 
\end{figure*}

\begin{figure*}[hbt!] 	
\centering 	
\includegraphics[scale=0.33]{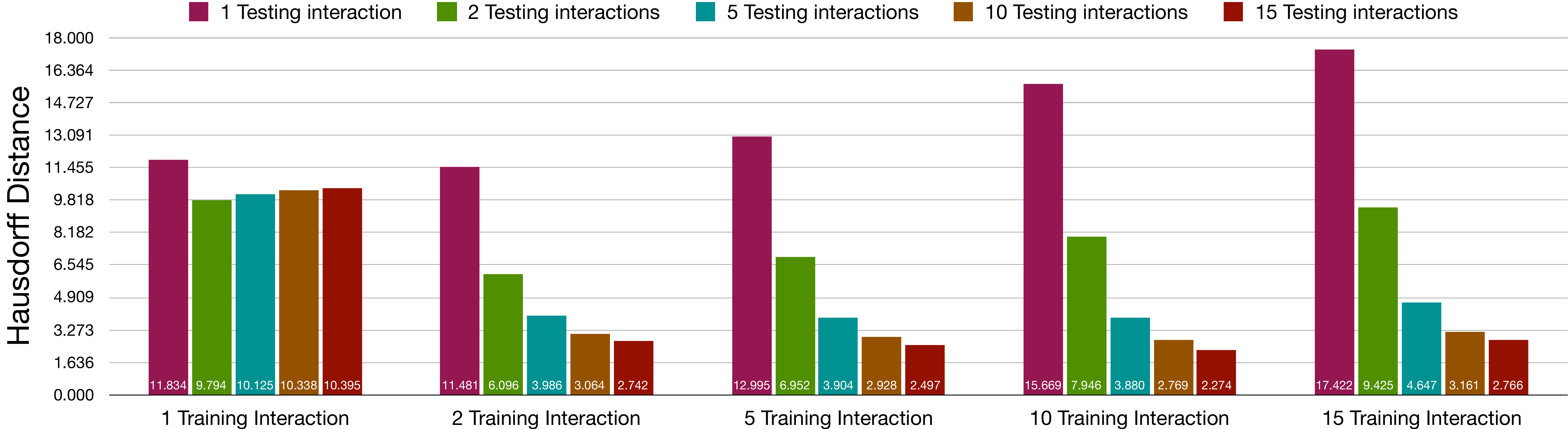} 	
\caption{Average Hausdorff Distance across cross-validation folds (lower is better). Different amounts of training-time and test-time interaction have been supplied to the algorithm. Groups represent the amount of interactions per slice and organ during training. Colors represent the amount of interactions during testing.} \label{fig:crossVHD} 
\end{figure*}

\begin{figure*}[hbt!]	
\centering 	
\includegraphics[scale=0.33]{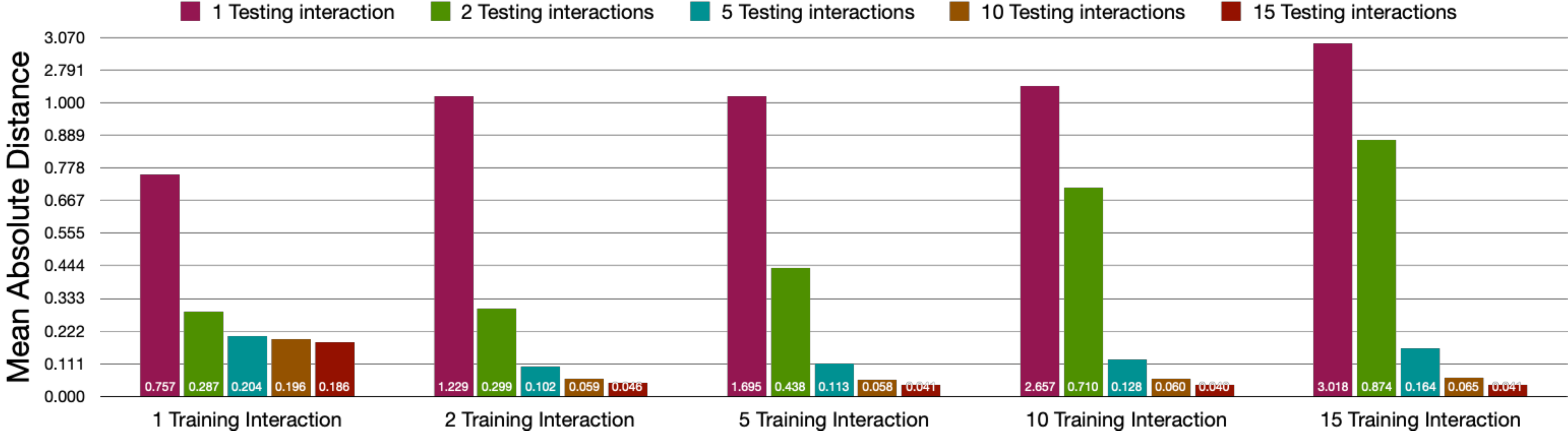} 	
\caption{Average Mean Absolute Distance (MAD) across cross-validation folds (lower is better). Different amounts of training-time and test-time interaction have been supplied to the algorithm. Groups represent the amount of interactions per slice and organ during training. Colors represent the amount of interactions during testing.} \label{fig:crossVMAD} 
\end{figure*}

To demonstrate the capabilities of our approach and draw conclusions and comparisons about our method, we perform four groups of experiments: i) 10-fold-cross validation with different amounts of interaction during both training and validation; ii) "known" region segmentation, where we evaluate the performance of our algorithm on a previously seen structure on a dataset disjoint from the one used for training (MSD: spleen); iii) "unknown" region segmentation, where we evaluate the generalization capabilities of our approach when segmenting a previously unseen region (MSD: colon cancer); iv) comparison to fully automatic methods on the test set of the "BCV" dataset.

Our approach has been trained from scratch using the same set of hyper-parameters. Please refer to Table \ref{table: hyperparams} for more detailed information. No training- or test-time augmentation has been performed. The data has been only re-sampled and standardized to zero-mean unit-standard deviation of the CT intensities (Table \ref{table: hyperparams}).


\begin{table}
\caption{Summary of hyperparameters used dduring training of our approach. }
\begin{centering}
\begin{tabular}{|c|c|c|}
\hline 
Parameter & Value & Notes\tabularnewline
\hline 
\hline 
Batch Size & $16$ & Multi-GPU Training\tabularnewline
\hline 
Learning Rate & $0.0001$ & Fixed throughout training\tabularnewline
\hline
Optimizer & Adam & \tabularnewline
\hline 
Random Seed & $4242$ & To reproduce the results\tabularnewline
\hline 
Max Interactions & $[1,2,5,10,15]$ & One value per experiment\tabularnewline
\hline 
Resolution & $[1.0,1.0]$ & Millimeters x and y direction\tabularnewline
\hline 
Size & $[512,512]$ & Pixels x and y direction\tabularnewline
\hline 
Click smoothing & sigma=$2$ & Gaussian smoothing for user clicks\tabularnewline
\hline

\end{tabular}
\par\end{centering}
\label{table: hyperparams}
\end{table}

\subsection{Experiments in cross validation}
\label{section:experiments:crossvalidation}
We perform 10-fold cross validation of our approach. We train our method on each of the 10 training sets and evaluate on the data left out from the training procedure during each experiment. We repeat this procedure 5 times, each time with a different amount of training-time interaction, as shown in Table \ref{table: hyperparams} (Max Interactions).

As a result we are able to report the performance of our approach using a variable amount of training and testing interactions. We simulate user interactions using the procedure explained in Section \ref{section:interaction}. 

The approach has been coded in python. The DL framework used for this work was PyTorch 1.0. Training was performed on multiple GPUs, each processing a portion of the batches (data parallelism). $4$ NVIDIA Tesla V100 GPUs with 16 GB of video memory have been used for each training. Training time was between 6 to 24 hours depending on the amount of training-time interaction. A total of $50$ training jobs have been run using a total of $200$ GPUs. Testing (in-house) has been performed on a single NVIDIA DGX Station with 4 Tesla V100 GPUs. 

After running our tests, we obtain the results reported in Figures \ref{fig:crossVDice}, \ref{fig:crossVHD} and \ref{fig:crossVMAD}. These metrics are computed slice-wise. We notice a few interesting aspects of the method: i) multiple interactions when creating training data improve results, especially when multiple interactions are supplied supplied during testing; ii) training with only one interaction does not result in an interactive network supporting multiple user clicks; iii) as the number of training-interactions increases, the performance for low amounts of test-time interaction drops; iv) as the number of test-time interaction increases, the network remains stable and delivers increasingly accurate results, regardless of the amount of (multiple) training-time interaction that was used.

From this experimental evidence it appears that there is a trade-off between the number of training interactions and the performance on the test set. In particular, depending on the task, it is necessary to carefully gauge the users needs and their willingness to spend time interacting with the network. That is, depending on the expected amount of test-time interaction different versions of the trained network are preferable. 

\subsection{Performance on a known region}
\label{section:experiments:kregions}

\begin{figure} 	
\centering 	
\includegraphics[scale=0.30]{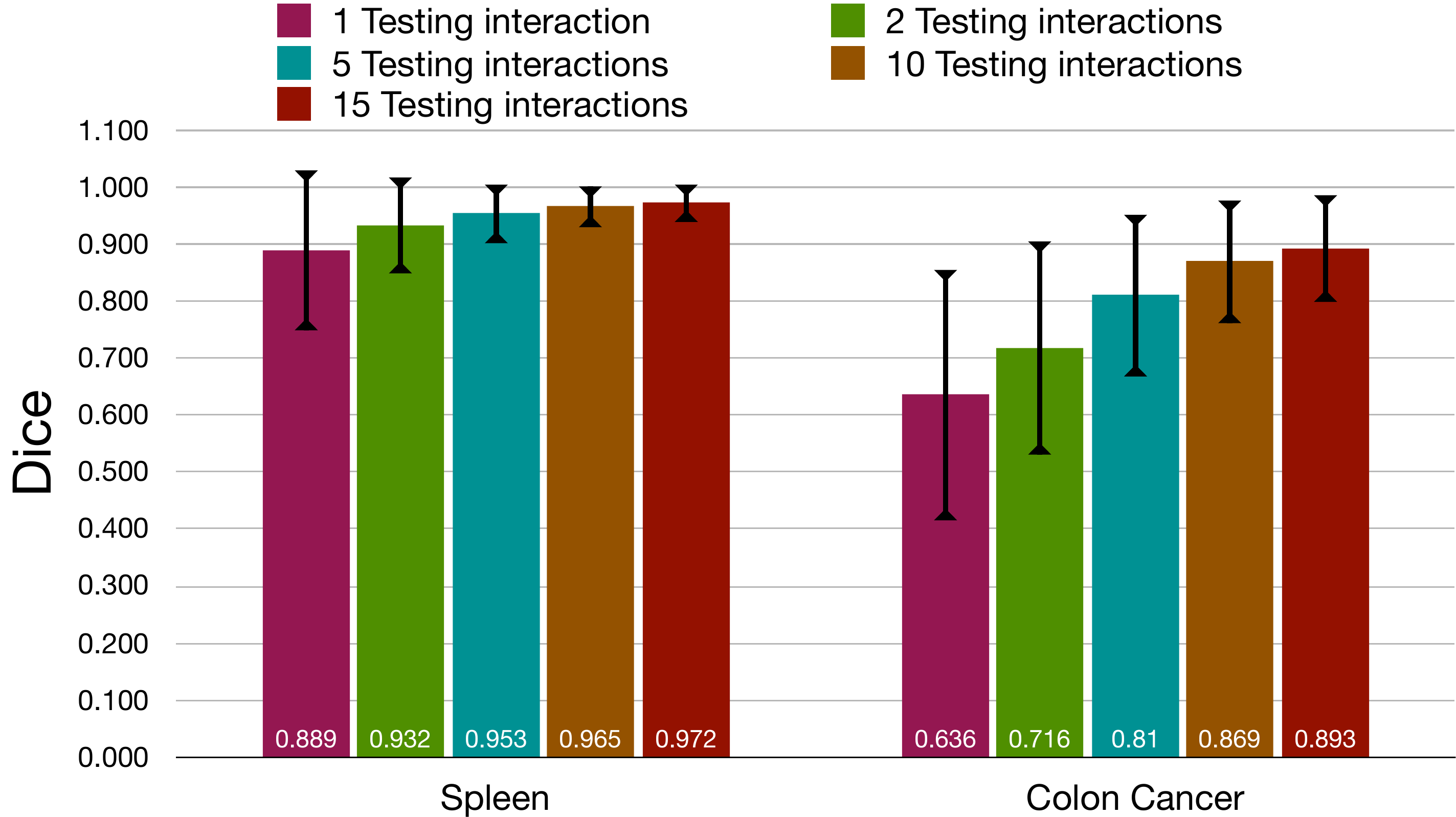} 	
\caption{Average Dice coefficients across cross-validation folds (higher is better), starting with a system trained to recognize structures annotated in BCV. The algorithm was never trained on MSD dataset. Each group of experiments corresponds to different segmentation tasks of the MSD challenge (horizontal axis). Performance corresponding to different amounts of test-time interaction is represented with different colors.} \label{fig:MSDice} 
\end{figure}

\begin{figure*} 	
\centering 	
\includegraphics[scale=0.50]{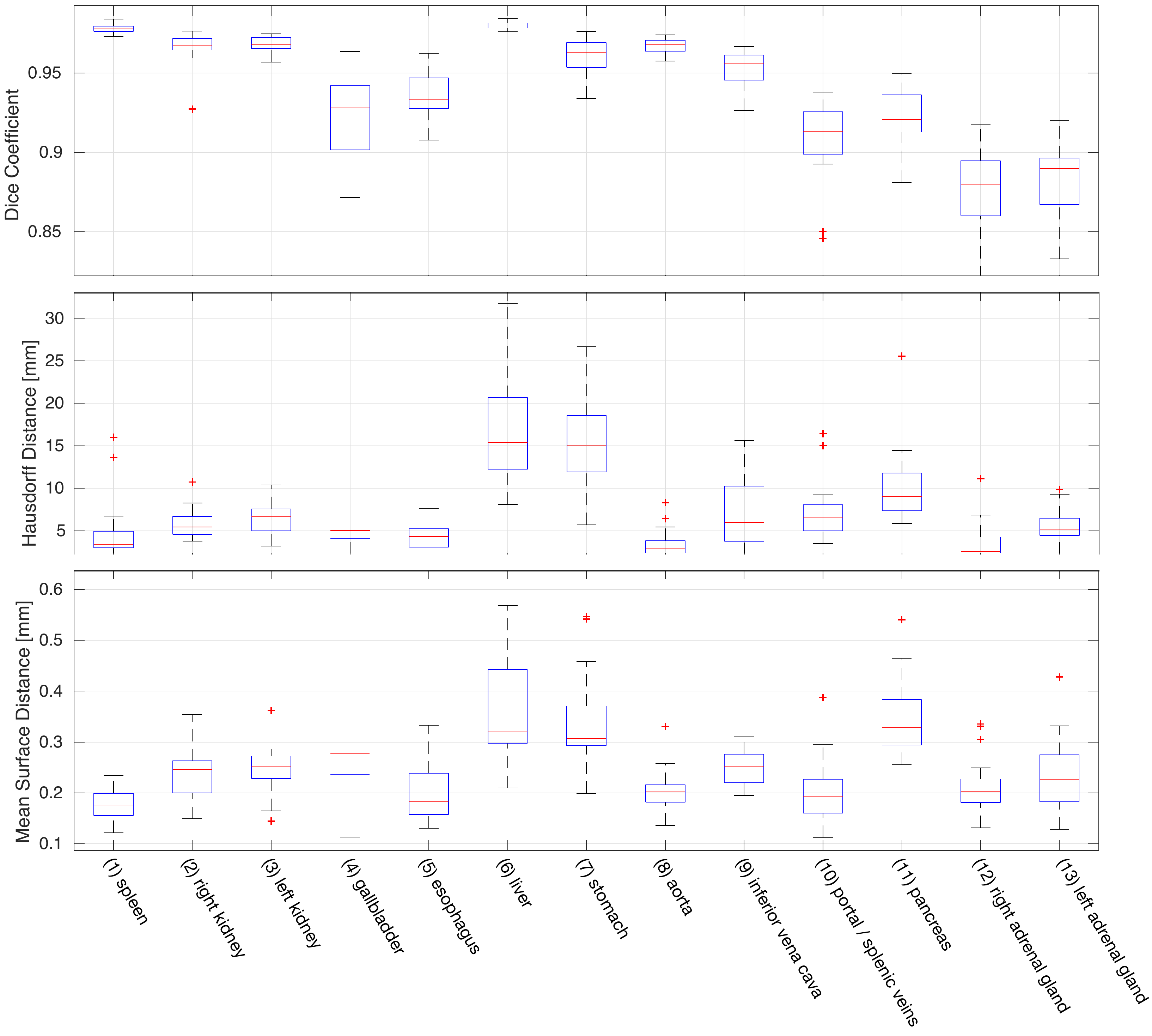} 	
\caption{Performance of our approach on the test set of "BCV" dataset. We show box plots for each organ indicating the median (red), the 25th and 75th percentiles (blue box), the minimum and maximum values (not including outliers, indicated by black horizontal bars), and outliers marked by red crosses. We thank the organizers of the "Multi-atlas labeling: beyond the cranial vault" MICCAI challenge for testing our method on their test dataset.} \label{fig:BCV} 
\end{figure*}

We demonstrate our algorithm on a previously unseen datasets from the Medical Segmentation Decathlon challenge. In this case, we aim to segment a "known" structure: the spleen. In order to simulate user interaction with the technique presented in \ref{section:interaction}, we need to make use of ground truth annotations. We therefore run our tests on publicly available annotated data downloaded from the MSD website. The dataset contains a total of $41$ annotated volumes. 

We use models trained with maximum $5$ interactions. This choice reflects our belief that these models deliver a good performance for a large interval of number of interactions. We repeat the experiment using each of the models resulting from 10-fold-cross validation, and report average performance in terms of Dice in Figure \ref{fig:MSDice}.

With little user interaction (2 clicks) the user already obtains high quality results, which improve as more interactive guidance is provided. These results match what was observed during 10-fold-cross validation and surpass state of the art for this task (Table \ref{table:MSD}).

\subsection{Performance on a unknown region}
\label{section:experiments:unkregions}

Our method has also been tested on a previously "unknown" structure from the MSD challenge dataset. The task we tackle is CT segmentation of colon cancer. We never trained our algorithm for this task, and therefore this experiment is crucial to give us some insights on the generalization capabilities of our FCNN when users wish to segment abnormalities or uncommon structures. We test on an annotated dataset that can be freely obtained from the MSD challenge website which contains $126$ annotated scans.

Similarly to the previous experiment, we use a model trained with a maximum of $5$ interactions. Results are reported for different levels of test-time interactions in Figure \ref{fig:MSDice}. We achieve higher Dice coefficients than a state-of-the-art method \cite{isensee2018nnu} trained specifically for this task (Table \ref{table:MSD}).

\subsection{Experiments on the test set}
\label{section:experiments:testset}

\begin{table}
\caption{Performance achieved by our algorithm on the MSD dataset in comparison
with state-of-the-art (SoA) results.}
\centering{}%
\begin{tabular}{|c|c|c|c|c|c|c|}
\hline 
Anatomy & SoA Dice & \multicolumn{5}{c|}{Our Dices}\tabularnewline
\hline 
Interactions & \textendash{} & 1 & 2 & 5 & 10 & 15\tabularnewline
\hline 
\hline 
Spleen & $0.96$ \cite{isensee2018nnu} & $0.89$ & $0.93$ & $0.95$ & $\mathbf{0.96}$ & $\mathbf{0.97}$\tabularnewline
\hline 
Colon cancer & $0.56$ \cite{isensee2018nnu} & $\mathbf{0.64}$ & $\mathbf{0.72}$ & $\mathbf{0.81}$ & $\mathbf{0.87}$ & $\mathbf{0.89}$\tabularnewline
\hline 
\end{tabular}
\label{table:MSD}
\end{table}

\begin{table}
\caption{Average results on the "BCV" test dataset. The columns "HDD" and "MAD" respectively contain the average Hausdorff and mean absolute distances achieved by each algorithm.}
\centering{}%
\begin{tabular}{|c|c|c|c|}
\hline 
Approach & Dice & HDD & MAD\tabularnewline
\hline 
\hline 
\textbf{Ours} & $\mathbf{0.94023}$ & $\mathbf{7.3693}$ & $\mathbf{0.25351}$\tabularnewline
\hline
\hline 
"nnUNet" & $0.88104$ & $17.2583$ & $1.3894$ \tabularnewline
\hline 
"ResNet101KL" & $0.84968$ & $18.468$ & $1.4504$ \tabularnewline
\hline 
DLTK (U-Net) \cite{pawlowski2017dltk} & 0.81535 & 62.8724  & 1.8613\tabularnewline
\hline
Auto Context \cite{roth2018multi} & $0.7824$ & $26.0951$ & $1.9362$\tabularnewline
\hline 
DeepSeg (FCN) \cite{larsson2016deepseg} & $0.76728$ & $27.3397$ & $2.9587$\tabularnewline
\hline
\end{tabular}
\label{table:BCV}
\end{table}

Finally, we have tested our method on the volumes reserved for testing on the "BCV" dataset. The ground truth annotation of these volumes is kept secret by the organizers of the challenge. In order to run our approach, our method has been run by the organizers of the "BCV" challenge on their own hardware using the strategy presented in Section \ref{section:interaction} to simulate five online user interactions per slice. 

The results of this test are shown in Figure \ref{fig:BCV}. Our approach delivers better results than any other algorithm that was officially evaluated by the challenge organizers \footnote{Leaderboard available at \url{https://www.synapse.org/#!Synapse:syn3193805/wiki/217785}}. This does not come as a surprise since our method is semi-automatic and is built with user-guidance in mind, while the leaderboard shows the results attained by fully automatic methods which cannot benefit from test-time interactions. Although we understand that this comparison is not completely fair, the final goal of medical image segmentation is to obtain accurate delineations enabling critical tasks and potentially influencing the life of the patients. We therefore believe that these results convey not only important information and insights about the capabilities of our approach, but also justify the need for semi-automatic approaches in medical image segmentation. 
In Table \ref{table:BCV} we show total results and we compare our approach with other state of the art methods that have been evaluated on the "BCV" dataset. 
We also report results for each region in comparison with the current best approach according to the BCV challenge website leader-board. These results are reported in terms of Dice, Hausdorff distance (HD) and Mean absolute distance (MAD) in to Table \ref{table:BCV-regionwise}. Results for each region have been reported in order and names have been abbreviated due to formatting.

\begin{table*}
\caption{Region-wise results obtained by our algorithm in comparison with the
current best result reported on the website of the BCV challenge
as of the 19th or March 2019.}
\centering{}%
\begin{tabular}{|c|c|c|c|c|c|c|c|c|c|c|c|c|c|c|}
\hline 
Method & Metric & Spln. & R-Kid. & L-Kid. & G.Blad & Esoph. & Liver & Stom. & Aorta & IVC & P/S V. & Pancr. & R-AG & L-AG\tabularnewline
\hline 
\hline 
\multirow{3}{*}{Ours} & Dice & \textbf{0.978} & \textbf{0.966} & \textbf{0.968} & \textbf{0.923} & \textbf{0.935} & \textbf{0.980} & \textbf{0.962} & \textbf{0.967} & \textbf{0.954} & \textbf{0.908} & \textbf{0.922} & \textbf{0.876} & \textbf{0.882}\tabularnewline
\cline{2-15} 
 & HDD & \textbf{4.785} & \textbf{5.848} & \textendash{} & \textendash{} & \textbf{4.303} & \textbf{16.582} & 15.\textbf{244} & \textbf{3.367} & \textbf{7.297} & \textbf{7.289} & \textbf{10.180} & \textbf{3.539} & \textbf{5.512}\tabularnewline
\cline{2-15} 
 & MAD & \textbf{0.177} & \textbf{0.239} & \textendash{} & \textendash{} & \textbf{0.200} & \textbf{0.356} & \textbf{0.343} & \textbf{0.205} & \textbf{0.249} & \textbf{0.2064} & \textbf{0.351} & \textbf{0.215} & \textbf{0.234}\tabularnewline
\hline 
\hline 
\multirow{3}{*}{BCV Best} & Dice & 0.968 & 0.921 & 0.957 & 0.783 & 0.834 & 0.975 & 0.920 & 0.925 & 0.870 & 0.836 & 0.830 & 0.788 & 0.781\tabularnewline
\cline{2-15} 
 & HDD & 6.489 & 10.88 & \textendash{} & \textendash{} & 10.31 & 28.50 & 30.38 & 19.98 & 19.74 & 23.41 & 17.74 & 6.58 & 7.41\tabularnewline
\cline{2-15} 
 & MAD & 0.370 & 0.831 & 0.470 & \textendash{} & 1.000 & 0.993 & 1.862 & 1.289 & 1.588 & 0.866 & 1.166 & 0.591 & 0.687\tabularnewline
\hline 
\end{tabular}
\label{table:BCV-regionwise}
\end{table*}

\section{Discussion and Concluding remarks}

We have presented a method to perform semi-automatic segmentation of medical data, in particular CT images, using a FCNN. In our approach, users can interact in real-time with the segmentation results by simply clicking on any structure that they want to segment. Any false-positive and false-negative regions can be corrected by placing any number of additional clicks. 

In addition to being able to segment any of the 13 regions observed during training, the same model has also generalized to structures that were not part of the training set. Testing our method to segment an unseen structure - colon cancers, it took only one mouse click per slice within the tumor to get a Dice score that is better than the best automated model that was specifically trained for the task (Dice 0.64 vs. 0.56). This result increased further up to Dice 0.81 provided 5 clicks and Dice 0.89 with 15 clicks. 

The choice of clicks as the form of interaction was for two important reasons. Clicks are a very simple form of interaction and allow for quick user interaction. Moreover, unlike scribbles, they are easy to simulate as they require only to locate the position of one pixel for interaction. Today's GPUs allow quick inference and our model takes only 0.04s to produce an output. Therefore, the user can see the results practically instantly. While holding the mouse button pressed and moving the mouse around, the user can also see how the result differs depending on where the click is finally placed. This further enables quick search for optimal click placement in practice. This was unaccounted for in testing, but it provides an additional mode of interaction that can additionally improve performance.

We anticipate that this approach can be generalized to other structures and imaging modalities. We are also excited to explore whether there is a limit to the number of structures that one model can learn to segment well provided simple mouse clicks. As areal and volume measurements are shown to be preferred over caliper-based measuring in radiology, we hope that our approach can be implemented as a tool for the radiologist in clinical practice to get accurate, quick and instantly validated segmentation results.

\section{Acknowledgements}
We would like to acknowledge Prof. Bennett A. Landman and Prof. Yuankai Huo for their time, help and efforts during evaluation of our method on the "Multi-atlas labeling: beyond the cranial vault" MICCAI challenge test dataset. 

\bibliographystyle{IEEEtran}
\bibliography{bibliography}

\begin{thebibliography}{10}
\providecommand{\url}[1]{#1}
\csname url@samestyle\endcsname
\providecommand{\newblock}{\relax}
\providecommand{\bibinfo}[2]{#2}
\providecommand{\BIBentrySTDinterwordspacing}{\spaceskip=0pt\relax}
\providecommand{\BIBentryALTinterwordstretchfactor}{4}
\providecommand{\BIBentryALTinterwordspacing}{\spaceskip=\fontdimen2\font plus
\BIBentryALTinterwordstretchfactor\fontdimen3\font minus
  \fontdimen4\font\relax}
\providecommand{\BIBforeignlanguage}[2]{{%
\expandafter\ifx\csname l@#1\endcsname\relax
\typeout{** WARNING: IEEEtran.bst: No hyphenation pattern has been}%
\typeout{** loaded for the language `#1'. Using the pattern for}%
\typeout{** the default language instead.}%
\else
\language=\csname l@#1\endcsname
\fi
#2}}
\providecommand{\BIBdecl}{\relax}
\BIBdecl

\bibitem{giger2008anniversary}
M.~L. Giger, H.-P. Chan, and J.~Boone, ``Anniversary paper: History and status
  of cad and quantitative image analysis: the role of medical physics and
  aapm,'' \emph{Medical physics}, vol.~35, no.~12, pp. 5799--5820, 2008.

\bibitem{iglesias2015atlas}
J.~E. Iglesias and M.~R. Sabuncu, ``Multi-atlas segmentation of biomedical
  images: A survey,'' \emph{Medical Image Analysis}, vol.~24, no.~1, pp. 205 --
  219, 2015.

\bibitem{peng2013graph}
B.~Peng, L.~Zhang, and D.~Zhang, ``A survey of graph theoretical approaches to
  image segmentation,'' \emph{Pattern Recognition}, vol.~46, no.~3, pp. 1020 --
  1038, 2013.

\bibitem{olabarriaga2001interactive}
S.~Olabarriaga and A.~Smeulders, ``Interaction in the segmentation of medical
  images: A survey,'' \emph{Medical Image Analysis}, vol.~5, no.~2, pp. 127 --
  142, 2001.

\bibitem{petitjean2011cardiac}
C.~Petitjean and J.-N. Dacher, ``A review of segmentation methods in short axis
  cardiac mr images,'' \emph{Medical Image Analysis}, vol.~15, no.~2, pp. 169
  -- 184, 2011.

\bibitem{nobel2006ultrasound}
J.~A. {Noble} and D.~{Boukerroui}, ``Ultrasound image segmentation: a survey,''
  \emph{IEEE Transactions on Medical Imaging}, vol.~25, no.~8, pp. 987--1010,
  Aug 2006.

\bibitem{foster2014pet}
B.~Foster, U.~Bagci, A.~Mansoor, Z.~Xu, and D.~J. Mollura, ``A review on
  segmentation of positron emission tomography images,'' \emph{Computers in
  Biology and Medicine}, vol.~50, pp. 76 -- 96, 2014.

\bibitem{leong2019segmentation}
C.~O. Leong, E.~Lim, L.~K. Tan, Y.~F. Abdul~Aziz, G.~S. Sridhar, D.~Socrates,
  K.~H. Chee, Z.-V. Lee, and Y.~M. Liew, ``Segmentation of left ventricle in
  late gadolinium enhanced mri through 2d-4d registration for infarct
  localization in 3d patient-specific left ventricular model,'' \emph{Magnetic
  resonance in medicine}, vol.~81, no.~2, pp. 1385--1398, 2019.

\bibitem{kline2017performance}
T.~L. Kline, P.~Korfiatis, M.~E. Edwards, J.~D. Blais, F.~S. Czerwiec, P.~C.
  Harris, B.~F. King, V.~E. Torres, and B.~J. Erickson, ``Performance of an
  artificial multi-observer deep neural network for fully automated
  segmentation of polycystic kidneys,'' \emph{Journal of digital imaging},
  vol.~30, no.~4, pp. 442--448, 2017.

\bibitem{ronneberger2015u}
O.~Ronneberger, P.~Fischer, and T.~Brox, ``U-net: Convolutional networks for
  biomedical image segmentation,'' in \emph{International Conference on Medical
  image computing and computer-assisted intervention}.\hskip 1em plus 0.5em
  minus 0.4em\relax Springer, 2015, pp. 234--241.

\bibitem{milletari2016v}
F.~Milletari, N.~Navab, and S.-A. Ahmadi, ``V-net: Fully convolutional neural
  networks for volumetric medical image segmentation,'' in \emph{3D Vision
  (3DV), 2016 Fourth International Conference on}.\hskip 1em plus 0.5em minus
  0.4em\relax IEEE, 2016, pp. 565--571.

\bibitem{cciccek20163d}
{\"O}.~{\c{C}}i{\c{c}}ek, A.~Abdulkadir, S.~S. Lienkamp, T.~Brox, and
  O.~Ronneberger, ``3d u-net: learning dense volumetric segmentation from
  sparse annotation,'' in \emph{International Conference on Medical Image
  Computing and Computer-Assisted Intervention}.\hskip 1em plus 0.5em minus
  0.4em\relax Springer, 2016, pp. 424--432.

\bibitem{an2017accuracy}
G.~An, L.~Hong, X.-B. Zhou, Q.~Yang, M.-Q. Li, and X.-Y. Tang, ``Accuracy and
  efficiency of computer-aided anatomical analysis using 3d visualization
  software based on semi-automated and automated segmentations,'' \emph{Annals
  of Anatomy-Anatomischer Anzeiger}, vol. 210, pp. 76--83, 2017.

\bibitem{itksnap}
P.~A. Yushkevich, J.~Piven, H.~Cody~Hazlett, R.~Gimpel~Smith, S.~Ho, J.~C. Gee,
  and G.~Gerig, ``User-guided {3D} active contour segmentation of anatomical
  structures: Significantly improved efficiency and reliability,''
  \emph{Neuroimage}, vol.~31, no.~3, pp. 1116--1128, 2006.

\bibitem{3dslicer}
A.~Fedorov, R.~Beichel, J.~Kalpathy-Cramer, J.~Finet, J.-C. Fillion-Robin,
  S.~Pujol, C.~Bauer, D.~Jennings, F.~Fennessy, M.~Sonka, J.~Buatti,
  S.~Aylward, J.~V. Miller, S.~Pieper, and R.~Kikinis, ``3d slicer as an image
  computing platform for the quantitative imaging network,'' \emph{Magnetic
  Resonance Imaging}, vol.~30, no.~9, pp. 1323 -- 1341, 2012.

\bibitem{kline2016semiautomated}
T.~L. Kline, M.~E. Edwards, P.~Korfiatis, Z.~Akkus, V.~E. Torres, and B.~J.
  Erickson, ``Semiautomated segmentation of polycystic kidneys in t2-weighted
  mr images,'' \emph{American Journal of Roentgenology}, vol. 207, no.~3, pp.
  605--613, 2016.

\bibitem{wang2018interactive}
G.~Wang, W.~Li, M.~A. Zuluaga, R.~Pratt, P.~A. Patel, M.~Aertsen, T.~Doel,
  A.~L. David, J.~Deprest, S.~Ourselin \emph{et~al.}, ``Interactive medical
  image segmentation using deep learning with image-specific fine-tuning,''
  \emph{IEEE Transactions on Medical Imaging}, 2018.

\bibitem{xu2016deep}
N.~Xu, B.~Price, S.~Cohen, J.~Yang, and T.~S. Huang, ``Deep interactive object
  selection,'' in \emph{Proceedings of the IEEE Conference on Computer Vision
  and Pattern Recognition}, 2016, pp. 373--381.

\bibitem{devaraj2017nodulevolume}
A.~Devaraj, B.~van Ginneken, A.~Nair, and D.~Baldwin, ``Use of volumetry for
  lung nodule management: Theory and practice,'' \emph{Radiology}, vol. 284,
  no.~3, pp. 630--644, 2017.

\bibitem{petitjean2011review}
C.~Petitjean and J.-N. Dacher, ``A review of segmentation methods in short axis
  cardiac mr images,'' \emph{Medical image analysis}, vol.~15, no.~2, pp.
  169--184, 2011.

\bibitem{udupa1996fuzzy}
J.~K. Udupa and S.~Samarasekera, ``Fuzzy connectedness and object definition:
  theory, algorithms, and applications in image segmentation,'' \emph{Graphical
  models and image processing}, vol.~58, no.~3, pp. 246--261, 1996.

\bibitem{boykov2006graph}
Y.~Boykov and G.~Funka-Lea, ``Graph cuts and efficient nd image segmentation,''
  \emph{International journal of computer vision}, vol.~70, no.~2, pp.
  109--131, 2006.

\bibitem{shi2000normalized}
J.~Shi and J.~Malik, ``Normalized cuts and image segmentation,''
  \emph{Departmental Papers (CIS)}, p. 107, 2000.

\bibitem{akkus2015semi}
Z.~Akkus, J.~Sedlar, L.~Coufalova, P.~Korfiatis, T.~L. Kline, J.~D. Warner,
  J.~Agrawal, and B.~J. Erickson, ``Semi-automated segmentation of
  pre-operative low grade gliomas in magnetic resonance imaging,'' \emph{Cancer
  Imaging}, vol.~15, no.~1, p.~12, 2015.

\bibitem{grady2005random}
L.~Grady, T.~Schiwietz, S.~Aharon, and R.~Westermann, ``Random walks for
  interactive organ segmentation in two and three dimensions: Implementation
  and validation,'' in \emph{International Conference on Medical Image
  Computing and Computer-Assisted Intervention}.\hskip 1em plus 0.5em minus
  0.4em\relax Springer, 2005, pp. 773--780.

\bibitem{menze2015multimodal}
B.~H. Menze, A.~Jakab, S.~Bauer, J.~Kalpathy-Cramer, K.~Farahani, J.~Kirby,
  Y.~Burren, N.~Porz, J.~Slotboom, R.~Wiest \emph{et~al.}, ``The multimodal
  brain tumor image segmentation benchmark (brats),'' \emph{IEEE transactions
  on medical imaging}, vol.~34, no.~10, pp. 1993--2024, 2015.

\bibitem{roth2018multi}
H.~R. Roth, C.~Shen, H.~Oda, T.~Sugino, M.~Oda, Y.~Hayashi, K.~Misawa, and
  K.~Mori, ``A multi-scale pyramid of 3d fully convolutional networks for
  abdominal multi-organ segmentation,'' in \emph{International Conference on
  Medical Image Computing and Computer-Assisted Intervention}.\hskip 1em plus
  0.5em minus 0.4em\relax Springer, 2018, pp. 417--425.

\bibitem{larsson2016deepseg}
M.~Larsson, Y.~Zhang, and F.~Kahl, ``Deepseg: Abdominal organ segmentation
  using deep convolutional neural networks,'' in \emph{Swedish Symposium on
  Image Analysis 2016}, 2016.

\bibitem{maninis2017deep}
K.-K. Maninis, S.~Caelles, J.~Pont-Tuset, and L.~Van~Gool, ``Deep extreme cut:
  From extreme points to object segmentation,'' \emph{arXiv preprint
  arXiv:1711.09081}, 2017.

\bibitem{agustsson2018interactive}
E.~Agustsson, J.~R. Uijlings, and V.~Ferrari, ``Interactive full image
  segmentation,'' \emph{arXiv preprint arXiv:1812.01888}, 2018.

\bibitem{castrejon2017annotating}
L.~Castrejon, K.~Kundu, R.~Urtasun, and S.~Fidler, ``Annotating object
  instances with a polygon-rnn.'' in \emph{Cvpr}, vol.~1, 2017, p.~2.

\bibitem{rother2004grabcut}
C.~Rother, V.~Kolmogorov, and A.~Blake, ``Grabcut: Interactive foreground
  extraction using iterated graph cuts,'' in \emph{ACM transactions on graphics
  (TOG)}, vol.~23, no.~3.\hskip 1em plus 0.5em minus 0.4em\relax ACM, 2004, pp.
  309--314.

\bibitem{wang2016slic}
G.~Wang, M.~A. Zuluaga, R.~Pratt, M.~Aertsen, T.~Doel, M.~Klusmann, A.~L.
  David, J.~Deprest, T.~Vercauteren, and S.~Ourselin, ``Slic-seg: A minimally
  interactive segmentation of the placenta from sparse and motion-corrupted
  fetal mri in multiple views,'' \emph{Medical image analysis}, vol.~34, pp.
  137--147, 2016.

\bibitem{wang2018deepigeos}
G.~Wang, M.~A. Zuluaga, W.~Li, R.~Pratt, P.~A. Patel, M.~Aertsen, T.~Doel,
  A.~L. Divid, J.~Deprest, S.~Ourselin \emph{et~al.}, ``Deepigeos: a deep
  interactive geodesic framework for medical image segmentation,'' \emph{IEEE
  Transactions on Pattern Analysis and Machine Intelligence}, 2018.

\bibitem{can2018learning}
Y.~B. Can, K.~Chaitanya, B.~Mustafa, L.~M. Koch, E.~Konukoglu, and C.~F.
  Baumgartner, ``Learning to segment medical images with scribble-supervision
  alone,'' in \emph{Deep Learning in Medical Image Analysis and Multimodal
  Learning for Clinical Decision Support}.\hskip 1em plus 0.5em minus
  0.4em\relax Springer, 2018, pp. 236--244.

\bibitem{simpson2019large}
A.~L. Simpson, M.~Antonelli, S.~Bakas, M.~Bilello, K.~Farahani, B.~van
  Ginneken, A.~Kopp-Schneider, B.~A. Landman, G.~Litjens, B.~Menze
  \emph{et~al.}, ``A large annotated medical image dataset for the development
  and evaluation of segmentation algorithms,'' \emph{arXiv preprint
  arXiv:1902.09063}, 2019.

\bibitem{isensee2018nnu}
F.~Isensee, J.~Petersen, A.~Klein, D.~Zimmerer, P.~F. Jaeger, S.~Kohl,
  J.~Wasserthal, G.~Koehler, T.~Norajitra, S.~Wirkert \emph{et~al.}, ``nnu-net:
  Self-adapting framework for u-net-based medical image segmentation,''
  \emph{arXiv preprint arXiv:1809.10486}, 2018.

\bibitem{pawlowski2017dltk}
N.~Pawlowski, S.~I. Ktena, M.~C. Lee, B.~Kainz, D.~Rueckert, B.~Glocker, and
  M.~Rajchl, ``Dltk: State of the art reference implementations for deep
  learning on medical images,'' \emph{arXiv preprint arXiv:1711.06853}, 2017.

\end{thebibliography}

\end{document}